\documentclass[lettersize,journal]{IEEEtran}
\usepackage{amsmath,amsfonts}
\usepackage{algorithmic}
\usepackage{algorithm}
\usepackage{array}
\usepackage[caption=false,font=normalsize,labelfont=sf,textfont=sf]{subfig}
\usepackage{textcomp}
\usepackage{stfloats}
\usepackage{url}
\usepackage{verbatim}
\usepackage{graphicx}
\usepackage{cite}
\usepackage{booktabs}
\usepackage{multirow}
\usepackage{color}

\def\BibTeX{{\rm B\kern-.05em{\sc i\kern-.025em b}\kern-.08em
    T\kern-.1667em\lower.7ex\hbox{E}\kern-.125emX}}
\usepackage{balance}

\begin{document}

\title{Amos-SLAM: An Anti-Dynamics Two-stage SLAM Approach}
\author{
\IEEEauthorblockN{
Yaoming~Zhuang$^{a,b*}$\thanks{This research was supported in part by the China Postdoctoral Science Foundation (2020M670778), the Natural Science Foundation of Liaoning Province (2021-BS-051), the Northeastern University Postdoctoral Research Fund (20200308), the Scientific Research Foundation of Liaoning Provincial Education Department (LJKQR20222509), the National Natural Science Foundation of China (U2013216, 61973093, 61901098, 61971118, and 61973063), the Fundamental Research Funds for the Central Universities (N2211004).}
,		Pengrun~Jia$^{a,b}$, 
		Zheng~Liu$^{a,b}$,
		Li~Li$^c$,
        Chengdong~Wu$^a$,
        Wei~Cui$^d$,
        ZhanLin~Liu$^e$
}\\
\IEEEauthorblockA{$^a$ Faculty~of~Robot~Science~and~Engineering,~Northeastern~University,~Shenyang~110819,~China;\\}
\IEEEauthorblockA{$^b$ College~of~Information~Science~and~Engineering,~Northeastern~University,~Shenyang~110819,~China;\\}
\IEEEauthorblockA{$^c$ JangHo~School~of~Architecture,~Northeastern~University,~Shenyang~110819,~China;\\}
\IEEEauthorblockA{$^d$ Institute~for~Infocomm~Research,~Astar~138632,~Singapore;\\}
\IEEEauthorblockA{$^e$ Microsoft,~Redmond,~Washington~98052,~USA;\\}
\IEEEauthorblockA{\IEEEauthorrefmark{1} Correspondence:~zhuangyaoming524@163.com}
}
\markboth{Journal of \LaTeX\ Class Files,~Vol.~14, No.~8, August~2021}%
{Shell \MakeLowercase{\textit{et al.}}: A Sample Article Using IEEEtran.cls for IEEE Journals}
\maketitle
\begin{abstract}

The traditional Simultaneous Localization And Mapping (SLAM) systems rely on the assumption of a static environment and fail to accurately estimate the system's location when dynamic objects are present in the background. While learning-based dynamic SLAM systems have difficulties in handling unknown moving objects, geometry-based methods have limited success in addressing the residual effects of unidentified dynamic objects on location estimation.
To address these issues, we propose an anti-dynamics two-stage SLAM approach. Firstly, the potential motion regions of both prior and non-prior dynamic objects are extracted and pose estimates for dynamic discrimination are quickly obtained using optical flow tracking and model generation methods. Secondly, dynamic points in each frame are removed through dynamic judgment. For non-prior dynamic objects, we present a approach that uses super-pixel extraction and geometric clustering to determine the potential motion regions based on color and geometric information in the image. Evaluations on multiple low and high dynamic sequences in a public RGB-D dataset show that our proposed method outperforms state-of-the-art dynamic SLAM methods.


\end{abstract}


\section{Introduction}

\IEEEPARstart{S}{imultaneous} Localization and Mapping (SLAM) aims at estimating the current pose and constructing a comprehensive global map for navigation, utilizing data acquired through sensors. It forms the foundation of robotics applications. Owing to advancements in the field of computer vision, vision-based SLAM systems have garnered significant attention from researchers in recent years. Currently, visual SLAM can be classified into monocular SLAM\cite{lsd, orb}, stereo SLAM\cite{dym, stereodso}, and RGB-D SLAM\cite{dense, Non} based on the type of camera utilized. Among these, RGB-D cameras have emerged as a popular choice among researchers, due to the availability of depth information and their robustness in indoor environments.


The existing visual SLAM systems perform well in many situations based on the static rigidity assumption\cite{survey}. However, the static environment hypothesis is often violated due to the presence of dynamic objects in the natural environment. Although some SLAM algorithms, such as ORB-SLAM2\cite{orb2}, can eliminate some dynamic features by using the RANSAC algorithm to classify a limited number of dynamic points as outliers of the static model, the algorithm becomes ineffective when there are a large number of dynamic objects in the background.


In recent years, advancements in computer vision and deep learning have led to the development of dynamic SLAM algorithms, such as DS-SLAM\cite{ds} and Dyna-SLAM\cite{dyna}, that deal with dynamic objects in the environment through semantic segmentation. These approaches have significantly improved the stability and accuracy of visual SLAM systems in dynamic environments as the dynamic objects can be detected and removed. However, the high computational cost of these segmentation algorithms poses a challenge to real-time applications in practice. Moreover, they can only handle known objects in the scene and fail to deal with unknown moving objects.


In order to effectively handle dynamic objects in natural environments, we propose a visual SLAM system that employs a two-stage dynamic segmentation algorithm. This algorithm divides dynamic targets into prior dynamic objects and unknown dynamic objects, and processes each type separately. 
Firstly, a real-time instance segmentation method is used to extract prior dynamic targets from the input image. Different from other methods (e.g., Towards Real-time(TR) SLAM\cite{tr}) only process Keyframes in pursuit of real-time, while Keyframes are not being updated. We process each frame to ensure that dynamic processing is not continuously affected by dynamic points. 
To address unknown dynamic objects in the scene, color and geometric information from the image is used to extract super-pixel blocks and perform geometric clustering. The potential motion regions are then extracted through reprojection errors of the clusters and the motion of the possible dynamic regions is determined through epipolar geometric constraints. Compared to traditional dynamic SLAM methods that perform motion judgment on a per-pixel basis, our proposal based on super-pixel blocks is more efficient and results in a more refined motion segmentation.


 The key contributions of this study can be summarized as follows:

\begin{enumerate}
\item In this paper, we proposed a real-time semantic RGB-D SLAM system based on ORB-SLAM2 for dynamic environments. 
\item{We proposed a two-stage dynamic detection method combining instance segmentation and visual geometry segmentation to better remove prior moving objects and unknown moving objects from the scene through improved motion judgment.}
\item{Several comparative experiments on the TUM dataset have shown that our method achieves the best accuracy compared to previous state-of-the-art dynamic SLAM methods.}
\end{enumerate}

\section{RELATED WORK}

Visual SLAM can be typically classified into two categories according to its front-end processing: the direct method and the feature method. To operate in dynamic environments, these methods employ different structures. The direct method often relies on geometric clustering and error analysis of image regions to distinguish dynamics, while the feature method utilizes geometric constraints to eliminate dynamic feature points.

\subsection{Direct methods of dynamic SLAM}
The main idea of these methods for dealing with dynamic objects is to divide the image into several clusters by space distance and to determine the dynamic cluster. Jaimez \textit{et al.} \cite{fast} perform a two-fold segmentation of the scene and track the scene flow to remove the dynamic in the background. Static Fusion(SF)\cite{sf} implements a dense reconstruction of the static environment by a dynamic probabilistic segmentation and an RGB-D fusion. Flow Fusion(FF)\cite{ff} introduces the optical flow and Scene Flow residuals on the basis of SF and applies them to dynamic segmentation and background reconstruction. Recently, DGFlow-SLAM\cite{dgf} divides the Scene Flow by a grid segmentation method and detects the dynamic object by an adaptive threshold method. It is worth noting that the Scene Flow is of great interest in many dynamic SLAM researches. It describes the perceived 3D motion field with respect to the camera and can be calculated by the dense Optical Flow and Ego Flow. Due to the computational cost of the dense Optical Flow and the error of Ego Flow, the Scene Flow can still be improved.
\subsection{Feature methods of dynamic SLAM}
The feature-based method of dynamic SLAM removes a prior object with the deep neural networks and non-prior objects with geometry constraints. DS-SLAM\cite{ds}only processes a prior object and determines the state of motion by checking the moving consistency. Dyna-SLAM\cite{dyna}combines instance segmentation and Multiview geometry to handle the dynamic scene. TR-SLAM\cite{tr} implements a real-time dynamic slam by semantic segmentation and reproject error. CFP-SLAM\cite{cfp} proposes a dynamic scene-oriented visual SLAM algorithm based on object detection and coarse-to-fine static probability. Although these methods propose many impressive ideas to the procession of SLAM in the dynamic scene, they can still be improved regarding location accuracy and real-time performance. Although a prior object can be addressed by the dynamic SLAM using selected deep neural networks, the issue of non-prior objects still has not been comprehensive addressed in the literature.  

\section{Methods}
\begin{figure*}[!t]
\centering
\includegraphics[width=4.5in]{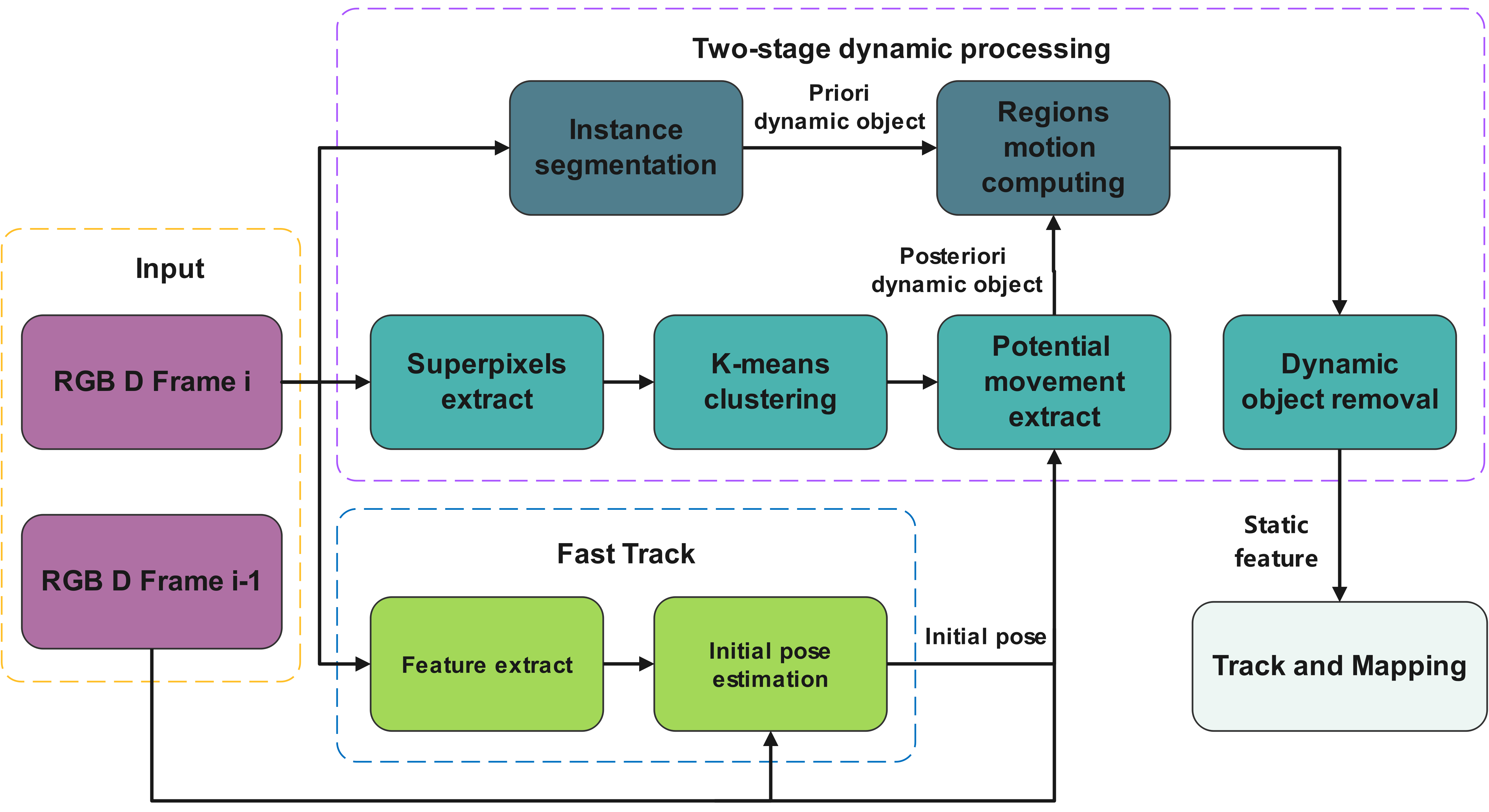}

\caption{The overall framework of our approach, with the dynamic processing within the purple dashed line, using adjacent frames to keep the motion estimates stable for real-time updates. After dynamic processing, the camera tracking is performed with static features.}
\label{fig1}
\end{figure*}

Classical SLAM systems are often based on static assumptions, while the natural environment contains many known and unknown dynamic moving objects. In an optimization-based approach, excessive dynamic feature points can lead to drift in the pose estimation of the SLAM system and render the location ineffective. Aiming to address the impact of the dynamic environment on the SLAM system, we propose a two-stage dynamic segmentation strategy that divides the dynamic segmentation problem into two stages: potential moving region acquisition and dynamic determination.
The specific procedure is demonstrated as follows: at the first stage, the potential moving regions are divided into prior dynamic objects and non-prior dynamic objects. The potential moving regions of a prior dynamic object can be obtained directly from instance segmentation. For non-prior dynamic objects, the scene is divided into different clusters through super-pixel clustering segmentation, and the potential dynamic regions are obtained by the average reprojection error. In the second stage, the actual dynamic parts are determined through improved motion judgment, and the dynamic objects in the scene are finally removed.
The overall framework of the system is shown in Fig \ref{fig1}.

\subsection{Instance segmentation module for handling prior dynamic objects}
We adopted YOLACT\cite{yolact} to process the prior objects. It is a fast-running real-time instance segmentation network that can accurately and continuously segment the contours of potential prior dynamic objects relative to current faster real-time target-detection algorithms, such as YOlOv5, and the lightweight semantic segmentation SegNet\cite{seg} used by DS-SLAM and TR SLAM. Compared to the Mask-RCNN\cite{mask} used by Dyna-slam, according to experiments in YOLACT, the Mask RCNN takes more than three times longer to process a frame than YOLACT under the same hardware conditions. Hence the real-time performance is far superior to that of Mask-RCNN. This module segments prior objects with high motion in the scene, mainly humans in indoor scenes, to the feature points that fall within the region of highly dynamic objects will be removed and will not be used for the map construction and tracking.

To improve the real-time performance of SLAM systems, most learning-based approaches, such as TR SLAM, only perform semantic segmentation on keyframes. Because of the better real-time performance of YOLACT, we perform segmentation on every frame. In addition, we perform instance segmentation using multi-threaded parallelism. It avoids the continuous impact of dynamic points until a new keyframe is extracted when the previous keyframe motion segmentation fails. The stability of the inter-frame pose estimation used for motion determination is guaranteed while considering real-time performance. Table \uppercase\expandafter{\romannumeral1} shows the comparison of different strategies.

\subsection{Visual geometry segmentation module for processing non-prior dynamic objects}
\begin{figure}[!t]
\centering
\subfloat[]{\includegraphics[width=1.1in]{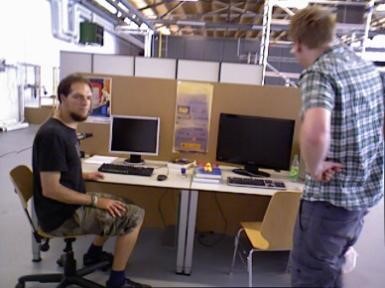}%
\label{a}}
\hfil
\subfloat[]{\includegraphics[width=1.1in]{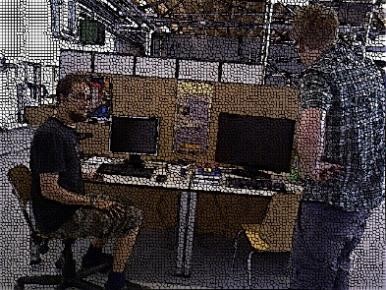}%
\label{fig_second_case}}
\hfil
\subfloat[]{\includegraphics[width=1.1in]{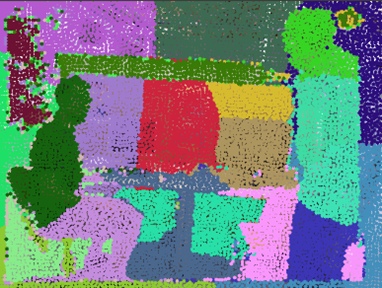}%
\label{fig_second_case}}
\caption{Visual geometry segmentation results. (a) is the input frames. (b) is the results of super-pixel extraction.(c) is the results of K-Mean clustering.}
\label{fig_sim}
\end{figure}

The semantic segmentation module can only remove fixed kinds of prior dynamic targets in the scene. It cannot handle objects that are affected by prior dynamic targets and change from static to dynamic or unknown moving objects. Unlike other methods that perform motion judgment pixel by pixel, we use a small number of super-pixel blocks instead of many pixels for dynamic processing, significantly reducing the memory burden on the system. More importantly, the outliers and noise in low-quality depth images can be reduced by extracting super-pixels.
The super-pixel was extracted from SLIC\cite{slic}. Clustering iterations are performed on r pixels around the initial seed point according to the clustering metric. The traditional SILC algorithm extracts super-pixel blocks based on color information and Euclidean distance, to which we add depth constraints for better segmentation of geometric boundaries. 
The clustering metric D for the super-pixel blocks is shown below:
\begin{equation}
\label{deqn_ex1a}
D = R(D_C)+R(D_E)+R(D_Z)
\end{equation}
\begin{equation}
\label{deqn_ex2a}
D_C= \sqrt{(L_i-L_c)^2+(A_i-A_c)^2+(B_i-B_c)^2}
\end{equation}
\begin{equation}
\label{deqn_ex3a}
D_E =\sqrt{(x_i-x_c)^2+(y_i-y_c)^2}
\end{equation}
\begin{equation}
\label{deqn_ex4a}
D_Z=\sqrt{d_i^2+d_c^2}
\end{equation}
where $D_C$, $D_E$, and$D_Z$ are the color metric, Euclidean distance metric, and depth metric respectively. $R()$ is the normalization function for comparing different elements. L, A, and B are the color channel values of the pixel in Lab space, respectively, and d is the depth value obtained in the depth image based on the pixel position. The subscripts i and c denote iteration process points and clustering centers, respectively.
After iteration, the set of super-pixel $S$ is obtained for each frame. Then the super-pixels in the set are clustered into m clusters by the k-means algorithm based on the spatial location of the super-pixel. Each super-pixel $S_i=[x_i,y_i,L_i,A_i,B_i,d_i,C_i]^T$ contains $x,y,L,A,B,d$ describes the average fetch value of the pixels in the region, and $C_i$ indicates the cluster label to which the super-pixel belongs after k-means clustering. The result of the visual geometry segmentation is shown in Fig \ref{fig_sim}. A rigid object may be divided into several neighboring clusters, with all the super-pixels in the cluster sharing motion constraints.

\subsection{Dynamic judgment of potential movement areas}

Our motion segmentation is divided into two main stages. The first stage is potential motion region extraction, mainly a judgment of geometric clustering as the potential motion regions of the prior dynamic targets can be obtained directly by instance segmentation. The second stage is the judgment of the motility of all potential motion regions.
In first stage, we choose the reprojection error as the condition for determining the potential motion region, the principle of which is shown in Fig 3.

It can be seen that reprojection error in dynamic is much greater than in static due to the effect of motion, so we set clusters with an average reprojection error above the threshold as potential dynamic clusters.
Error calculations are performed using temporary local map points in adjacent frames and detected corner points. The corner points are extracted from the last frame and tracked through the LK optical flow to get the matching points for the current frame. A 2D-3D inverse projection of the corner points of the previous frame is performed to obtain the temporary map points. The error is calculated for the temporary map points and the matching points according to the initial poses obtained from the tracking thread. The matching points with valid depths are aggregated according to the clustering information to get the average reprojection error of the geometric clusters and determine the potential motion region.

To obtain the reprojection error, we perform the calculation in (5) for each matched point pair with a valid depth.
\begin{equation}
\label{deqn_ex1a}
R_{{pe}_j} = \frac{1}{m}\sum_{i}^{m}\rho(||u_i-\pi(T_{wc}P_i)||^2) 
\end{equation}
where m is the number of matched points in the cluster, $\pi(\bullet)$ is the projection function for projecting 3D points onto the 2D pixel plane according to the camera model.$\rho(\bullet)$is a Cauchy robust function for outliers, u is the pixel coordinates of the matching points in the current frame, $T_{wc}$ is the pose transformation from the camera coordinate system to the world coordinate system , $P_i$ is the provisional map points in the world coordinate obtained by 2D-3D inverse projection of the matching points in the previous frame, and $P_i$ is calculated as follows:
\begin{figure}[!t]
\centering
\includegraphics[width=2in]{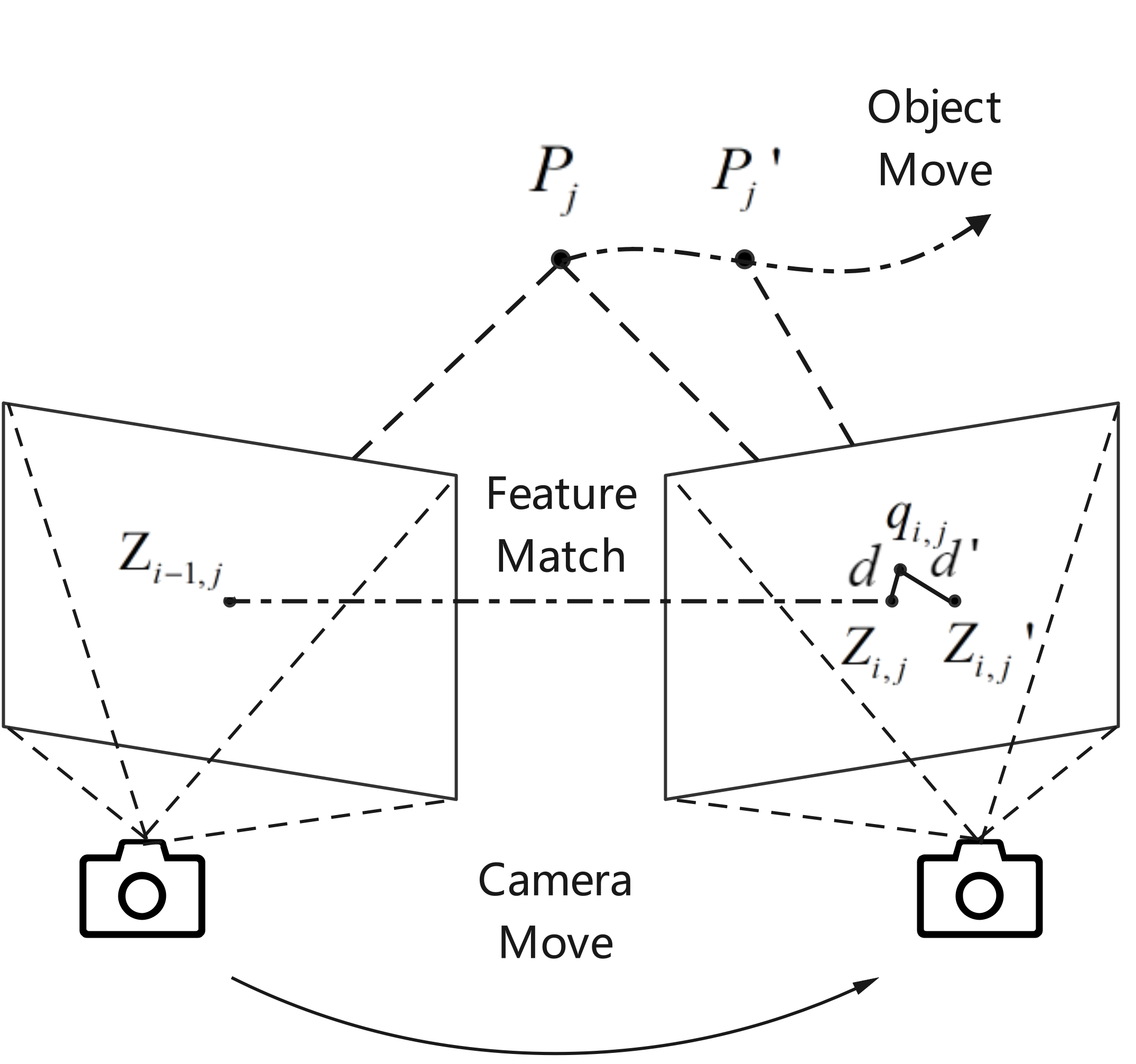}
\caption{Feature matching in a dynamic environment. $P_j$  is the position of the 3D point at the moment $j$. If 3D point $P$ is a static point, the projection point of point $P$ in the current frame is $z_{i,j}$. If $P$ is a dynamic point, the projection is $z_{i,j}'$. The match point $q_{i,j}$ is calculated by feature matching in the current frame. $d$ is the reprojection error when the point is static, and $d'$ is the reprojection error when the point is dynamic.}
\label{fig3}
\end{figure}
\begin{equation}
\label{deqn_ex1a}
P_i=T_{wl}[ Z\frac{u_{i-1}-c_x}{f_x},Z\frac{v_{i-1}-c_y}{f_y},Z]^T
\end{equation}
where $T_{wl}$ is the inverse transformation of pose estimate in the last frame, $[u_{i-1},v_{i-1}]$ is the pixel coordinates of the matching point in frame i-1, $c_x,c_y,f_x,f_y$ is the internal reference coefficient of the camera model, and Z is the depth of the matching point.

Because of feature mismatching or incomplete dynamic point removal, there are errors in pose estimation, leading to the possibility that the average reprojection error of some static clusters may also exceed the threshold. Refer to DS-SLAM for further use of epipolar constraints to determine the motion of potential dynamic regions. The specific implementation is as follows:

Firstly, the fundamental matrix is calculated according to the matched points, and then the polar line in the current frame get by the fundamental matrix. The distance between the matching point and the polar line is the constraint. If it exceeds the threshold, it is considered dynamic. If a point does not satisfy the epipolar constraint and falls within the potential motion region, the region is considered dynamic. At the same time, remove the super-pixel blocks close to the image boundary.
The polar line L for the current frame is calculated as:
\begin{equation}
\label{deqn_ex1a}
L=\begin{bmatrix} 
A\\B\\C 
\end{bmatrix}
=Fp_1=K^{-T}t  ^\wedge K^{-1}\begin{bmatrix} 
u\\v\\1 
\end{bmatrix}
\end{equation}
where A, B, and C are the coefficients of the 2D linear equations, F is the fundamental matrix, K is the internal reference matrix of the camera, $t^\wedge$ is the antisymmetric matrix of the translation vectors and $p_1=[u,v,1]^T$ is the normalized plane coordinates of the matched points in the last frame.

Then the distance from the matching point to the poles is:
\begin{equation}
\label{deqn_ex1a}
d_E=\frac{|p_2^TFp_1|}{\sqrt{||A||^2+||B||^2}}
\end{equation}

As the dynamic points have not been completely removed in the calculation of the fundamental matrix. To get a more accurate fundamental matrix, we use the distance from the matching point to the epipolar line to exclude some points that do not fit the constraint before proceeding to the second fundamental matrix calculation.

Through experiments, we found that the dynamic points detected using the epipolar constraints usually fall at the boundaries of dynamic objects, Determining the dynamics of the clusters directly based on where the dynamic points fall can lead to erroneous results when the pose estimates drift. For the boundary problem of prior dynamic targets, we perform morphological expansion of the target mask obtained from instance segmentation to expand the original mask range. For the boundary problem of non-prior dynamic objects, we use an iterative approach to compare 30 pixels around the dynamic points. The cluster will be set as dynamic when the average reproject error of the cluster to which pixel belong is the largest.

\begin{figure*}[!t]
\centering
\includegraphics[width=5in]{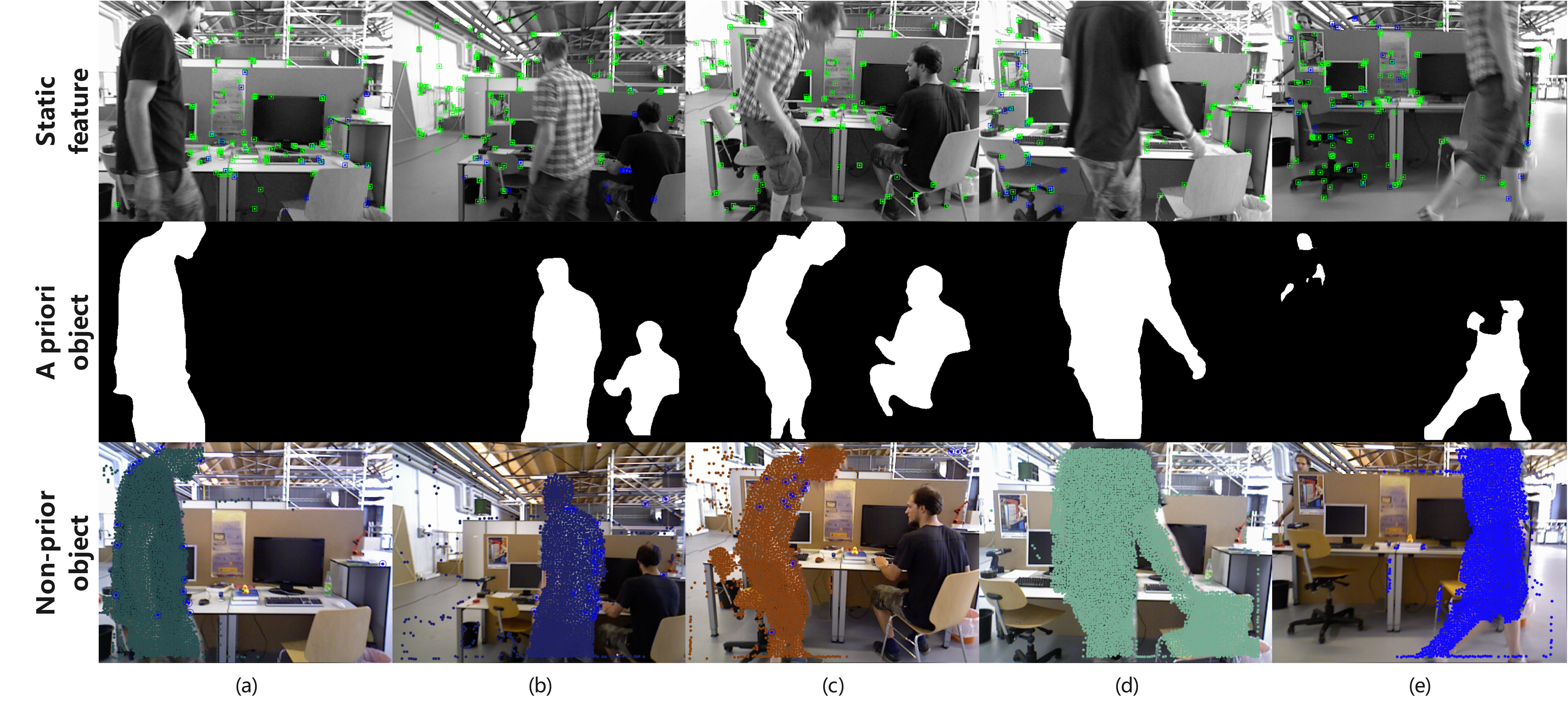}
\caption{Results of motion segmentation. The first row shows the feature point extraction after dynamic segmentation, the second row shows the a prior dynamic target obtained by instance segmentation, and the third row shows the dynamic target obtained by joint visual geometric segmentation. (a)-(e) are comparisons of the segmentation effects in different cases, respectively. Frame selected from sequence fr3/walking\_xyz}
\label{fig4}
\end{figure*}

The results of the segmentation of dynamic targets using our proposed method are shown in Fig 4. (a) shows that the proposed method can efficiently segment dynamic objects in the scene. The result (b) shows that the visual geometry module correctly segments the actual walking person in the case of a part of the prior dynamic target that is static while the instance segmentation removes both the walking and sitting person. In (c) and (d), as the person pushes the chair, it is transformed from static to dynamic. The instance segmentation cannot handle such prior static objects, which the visual geometry module successfully separates from the scene. (e), on the other hand, is a case of partial segmentation failure due to blurring caused by rapid motion.
\subsection{POSE ESTIMATION}

To ensure robust estimation, we propose a model generation method for initialization. Specifically, this method generates three models and calculates the reprojection error of each point according to the motion model. The points which are less than the threshold are regarded as the inlier of the model. the model with the most inlier is selected as the initial pose for the dynamic estimation.

The first model uses the feature points in the reference keyframe to match the current frame by the LK optical flow. After obtaining the matched points, the EPNP algorithm with RANSAC is used to compute the pose and inlier from the local map points and matched points; In the second model, to achieve fast pose estimation, we detect a sparse set of corner features from the last frame and track them. Because of the way of processing adjacent frames to make the feature matching and pose calculation as above, this model performs well when the reference keyframe is selected a long time from the current frame or when the present motion is too fast; The third model is obtained from the uniform motion model of the camera. It is very effective when the pose estimation is drifted. The initial pose estimates were then used for the dynamic division. After removing the dynamic points, the ORB-SLAM2 tracking process was performed.


\section{EXPERIMENTS AND RESULTS}
\subsection{Overview of the experiment}
\subsubsection{Overview of the experiment}
The TUM public dataset is widely used to evaluate RGB-D SLAM methods, and many dynamic SLAM methods have been tested on this dataset. To compare our proposed method cross-sectionally, we choose the TUM dataset to compare experimental effects. The trajectory compared with the ground truth is shown in the Fig 5. Then, an experimental comparison between processing only keyframes and all frames is performed to verify the impact of both strategies on accuracy. Finally, our results are compared with the current state-of-the-art dynamic SLAM methods, both direct and feature. The experimental results demonstrate that our method achieves optimal accuracy in multiple sequences. All experiments were implemented on a laptop with Intel(R) CPU i5-11400H @3.50GHz x12,16 GiB System memory and GeForce GTX 3060 GPU. For calculating the evaluation metrics, we chose the trajectory evaluation tool EVO, an open-source library for processing, evaluating, and comparing the trajectory output of odometry and SLAM algorithms.
\subsubsection{Dataset description}
TUM dataset: To demonstrate performance in a dynamic environment two low and four high dynamic sequences were selected for evaluation in the TUM dataset. fr3/sitting (fr3/s for short) sequence describes a low dynamic scene in which the moving part of the scene is the slight body movements of two people sitting in chairs. fr3/walking (fr3/ w) sequence is highly dynamic, and the dynamic objects in the scene are mainly two people in motion and something that move briefly under the influence of the people. Most algorithms that don’t have a procession of dynamic objects will fail entirely on the walking sequence. As the sequence name indicates, there are four types of camera movement, remaining almost static (static), moving in three directions (xyz), rotating along the main axis (rpy) and moving along a halfsphere with a diameter of one meter (half).
\subsubsection{Assessment indicators}
The error metrics used for evaluation are Absolute Trajectory Error (ATE) and Relative Pose Error (RPE), commonly used in SLAM. ATE measures the overall consistency of the trajectory, and RPE measures the odometer drift per second. Absolute trajectory error is described in units of (m/s), and relative pose error describes translational drift in (m/s) and rotational drift in (°/s), respectively. There are many ways to describe an evaluation metric, such as mean error, median error, mean square error, standard deviation, etc. Here we have chosen to use the Root Mean Square Error (RMSE) to compare the errors.


\begin{figure}[!t]
\centering
\includegraphics[width=3.5in]{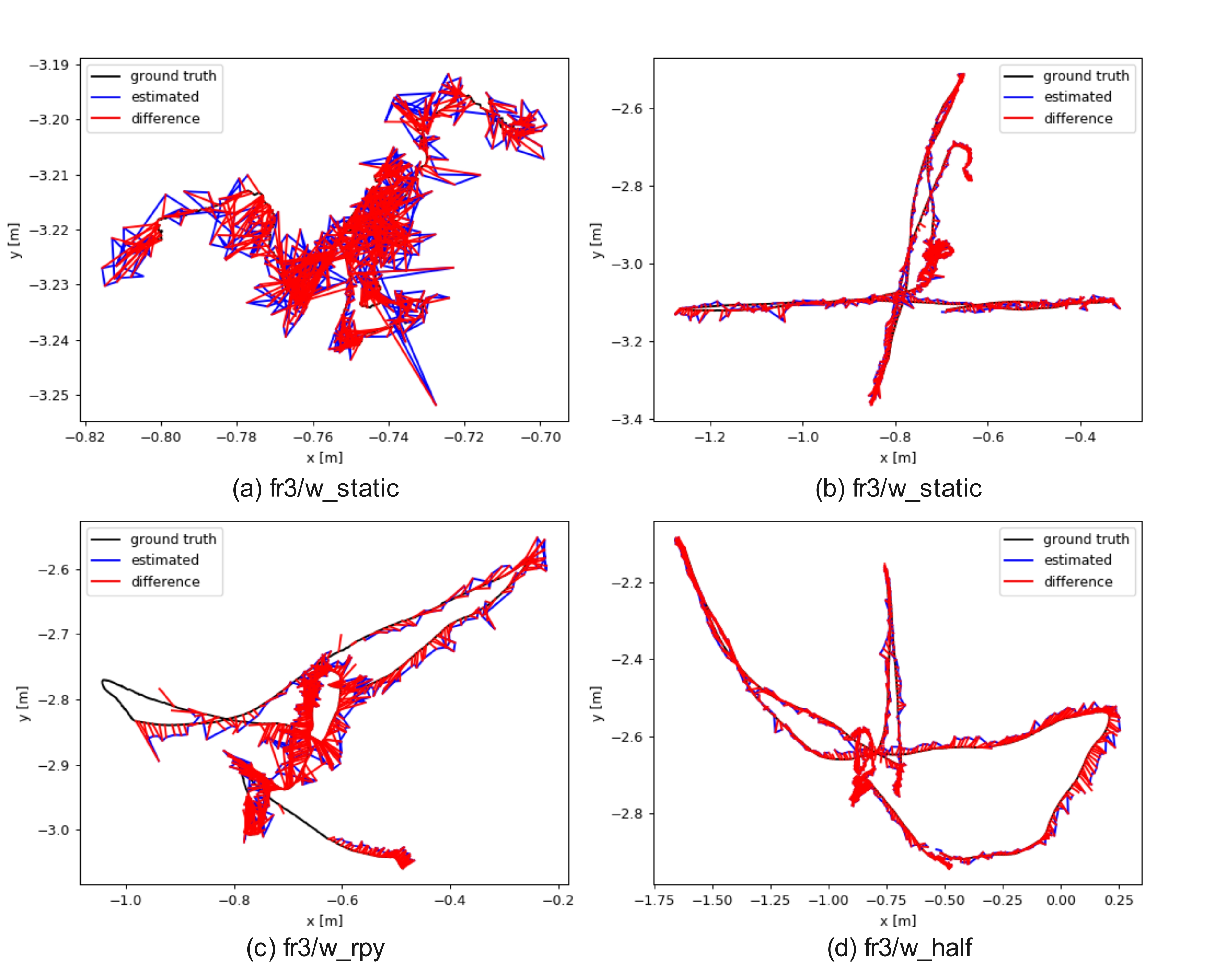}
\caption{Trajectory of Ours and Ground Truth}
\label{fig4}
\end{figure}

\subsection{Only-Keyframe versus full processing}

For the Keyframe selection strategy, we followed the original ORB-SLAM2 selection approach. The localization accuracy of our method using each strategy was tested for each of the four sequences of the highly dynamic scene while keeping other experimental conditions constant. Since the fully processed approach reduces the number of dynamic map points in the map to a greater extent, the localization accuracy is higher than that of the Only-KF. The experimental results in Table \uppercase\expandafter{\romannumeral1}also validate our conclusions.

\begin{table}[!t]
	\centering
	\caption{ Comparison of accuracy between OnlyKF and All}
	\begin{tabular}{lcccccc}
	\toprule
	\multirow{2}{*}{Sequence} & \multicolumn{2}{c}{ATE(m/s)} &\multicolumn{2}{c}{RTE(m/s)}&\multicolumn{2}{c}{RTE($^\circ$/s)} \\
	\cmidrule(lr){2-3} \cmidrule(lr){4-5}  \cmidrule(lr){6-7} 
	&OnlyKF &ALL &OnlyKF &ALL &OnlyKF  &ALL\\

	\midrule
	w\_static& 0.011 &		\textbf{0.007}&		0.012&\textbf{0.006} & 	0.269& 		\textbf{0.175}  \\
	w\_xyz& 	0.020 &		\textbf{0.013}&	 	0.016&\textbf{0.011}&		\textbf{0.322}& 0.386  \\
	w\_half& 	0.031 &		\textbf{0.027}& 	0.022&\textbf{0.020}& 	0.540& 		\textbf{0.508}  \\
	w\_rpy& 	0.035	&		\textbf{0.025}&		0.018 &\textbf{0.012}&	0.468&		\textbf{0.391}	\\
	\bottomrule
	\end{tabular}
\end{table}

\subsection{Comparision of accuracy with current SOTA papers}

Our experimental results are compared with the dynamic direct SLAM methods, Such as Static Fusion\cite{sf}, DGFlow-SLAM\cite{dgf} mainly based on geometric clustering, and the dynamic feature SLAM methods, Such as DS-SLAM\cite{ds}, Dyna-SLAM\cite{dyna}, TR-SLAM\cite{tr}, CFP-SLAM\cite{cfp} which based on learning and geometric constraints. Tables \uppercase\expandafter{\romannumeral2}-\uppercase\expandafter{\romannumeral4} show the accuracy of the dynamic SLAM systems for localization with different evaluation metrics, respectively. The best results are shown in bold and the next best results are underlined. The "/" means data lost and all data get from the results published in the original paper. Table \uppercase\expandafter{\romannumeral2} describes the overall consistency of the trajectory, Table \uppercase\expandafter{\romannumeral3} describes the translational drift of the odometer per second, and Table \uppercase\expandafter{\romannumeral4} describes the rotational drift of the odometer per second. 

By analyzing the data in the table, we note that all methods using optical flow have relatively large errors in the rpy sequences. We think it is partly due to the light variations in a portion of the rpy sequence, which breaks the Grayscale invariant hypothesis required for optical flow calculations. Although the overall accuracy errors of the dynamic SLAM methods based on geometric clustering are higher than those of the learning-based methods, these methods do not require prior information about moving objects and will outperform the learning-based methods in the face of unknown dynamic objects. As our method achieves the best motion segmentation of dynamic scenes, we achieve optimal accuracy in multiple metrics across multiple sequences. The experimental results also demonstrate that our method is the most accurate dynamic SLAM method for handling dynamic indoor environments.
\begin{table}[!t]
	\centering
	\caption{RMSE comparison of absolute trajectory errors ATE (m/s)}
	\begin{tabular}{lccccccc}
	\toprule
	\multirow{2}{*}{Sequence} & \multicolumn{2}{c}{Direct SLAM} &\multicolumn{5}{c}{Feature SLAM} \\
	\cmidrule(lr){2-3} \cmidrule(lr){4-8} 
	&SF &DGF &DS &Dyna &TR  &CFP &Ours \\

	\midrule

	s\_xyz& 	0.040 &	/&			/& 		0.015 & 		0.011 & \underline{0.009} & \textbf{0.008} \\
	s\_half& 	0.040 &	/&			/&\underline{0.017} &\underline{0.017}& \textbf{0.014} & \textbf{0.014} \\
	w\_static& 	0.014 &	0.008&		0.008&\textbf{0.006} & 	0.011& \textbf{0.006} & \underline{0.007} \\
	w\_xyz& 	0.127 &	0.028&	 	0.024&\underline{0.015}&	0.019& \textbf{0.014} & \textbf{0.014} \\
	w\_half& 	0.391 &	0.048& 		0.030&\underline{0.025}& 	0.029& \textbf{0.023} & \underline{0.025} \\
	w\_rpy& 	/	&\textbf{0.027}&	0.444 &\underline{0.035}&	0.037&		0.036 & \textbf{0.027} \\
	\bottomrule
	\end{tabular}
\end{table}
\begin{table}[!t]
	\centering
	\caption{RMSE comparison of relative positional errors RPE (m/s)}
	\begin{tabular}{lccccccc}
	\toprule
	\multirow{2}{*}{Sequence} & \multicolumn{2}{c}{Direct SLAM} &\multicolumn{5}{c}{Feature SLAM} \\
	\cmidrule(lr){2-3} \cmidrule(lr){4-8} 
	&SF &DGF &DS &Dyna &TR  &CFP &Ours \\

	\midrule

	s\_xyz& 	0.028 &	/&			/& 		0.014 & 		0.016 & \underline{0.011} & \textbf{0.007} \\
	s\_half& 	0.030 &	/&			/&\underline{0.016} &		0.025& \underline{0.016} & \textbf{0.008} \\
	w\_static& 0.013 &	0.011&		0.010&\underline{0.008} & 	0.011& \underline{0.008} & \textbf{0.006} \\
	w\_xyz& 	0.121 &	0.039&	 	0.033&			0.021&	0.023& \underline{0.019} & \textbf{0.011} \\
	w\_half& 	0.207 &	0.044& 		0.029&			0.028& 	0.042& \underline{0.025} & \textbf{0.012} \\
	w\_rpy& 	/	&	0.197&		0.150 &\underline{0.044}&	0.047&		0.050 & \textbf{0.020} \\
	\bottomrule
	\end{tabular}
\end{table}
\begin{table}[!t]
	\centering
	\caption{RMSE comparison of relative positional error RPE (°/s)}
	\begin{tabular}{lccccccc}
	\toprule
	\multirow{2}{*}{Sequence} & \multicolumn{2}{c}{Direct SLAM} &\multicolumn{5}{c}{Feature SLAM} \\
	\cmidrule(lr){2-3} \cmidrule(lr){4-8} 
	&SF &DGF &DS &Dyna &TR  &CFP &Ours \\

	\midrule

	s\_xyz& 	0.920		&	/&/			&0.504 & 		0.596 & \underline{0.487} & \textbf{0.312} \\
	s\_half& 	2.110 	&	/&/			&0.704 &		{0.789}& \underline{0.591} & \textbf{0.367} \\
	w\_static& 0.380 	&0.280&0.269		&0.261& 0.287& \underline{0.252} & \textbf{0.175} \\
	w\_xyz& 	2.660 	&0.618&0.826		&0.628&	0.636& \underline{0.602} & \textbf{0.386} \\
	w\_half& 	5.040 	&1.069&0.814		&0.784& 	0.965& \underline{0.757} & \textbf{0.391} \\
	w\_rpy& 	/		&3.728&3.004 		&\underline{0.989}&	1.058&		1.108 & \textbf{0.508} \\
	\bottomrule
	\end{tabular}
\end{table}
\section{Conclusion}
This paper presents Amos-SLAM, an anti-dynamics two-stage SLAM approach based on vision and geometry. The proposed Amos-SLAM leverages instance segmentation to handle prior dynamic objects and a combination of super-pixel and k-means clustering to address unknown dynamic objects in the environment. Our approach processes all frames to improve the performance and results show it achieves an improved localization accuracy compared to current state-of-the-art techniques.

Future work will focus on the improvement of dynamic object recognition by using Surfel representation and semantic dense mapping of the static environment. Additionally, more accurate dynamic judgment methods will be explored to further enhance the performance of our system.



\bibliographystyle{IEEEtran}
\bibliography{refs}

\vfill

\end{document}